\documentclass{article}

\PassOptionsToPackage{table}{xcolor}
\usepackage{iclr2027/iclr2027_conference,times}
\iclrfinalcopy 

\usepackage[utf8]{inputenc} 
\usepackage[T1]{fontenc}    
\usepackage{hyperref}       
\usepackage{url}            
\usepackage{booktabs}       
\usepackage{amsfonts}       
\usepackage{nicefrac}       
\usepackage{microtype}      
\usepackage{graphicx}
\usepackage{pifont}
\usepackage{enumitem}
\usepackage{multirow}
\usepackage{makecell}
\usepackage{placeins}
\usepackage{xcolor}
\usepackage{bm}
\newcommand{\cmark}{\ding{51}}
\newcommand{\xmark}{\ding{55}}

\newcommand{\model}{CompactionRL}

\usepackage[most]{tcolorbox}
\usepackage{fvextra}
\newtcolorbox{promptbox}[1][]{
  breakable,
  colback=gray!4,
  colframe=gray!45,
  boxrule=0.5pt,
  arc=1mm,
  left=5pt,
  right=5pt,
  top=5pt,
  bottom=5pt,
  before skip=6pt,
  after skip=6pt,
  title=#1,
  fonttitle=\bfseries\small,
}

\title{\model: Reinforcement Learning with Context Compaction for Long-Horizon Agents}

\author{
  Yujiang Li\textsuperscript{*$^\dagger$} \quad
  Zhenyu Hou\textsuperscript{*$^\dagger$} \quad
  Yi Jing\textsuperscript{$^\dagger$} \quad
  Jie Tang\textsuperscript{} \quad  
  Yuxiao Dong\textsuperscript{} 
  \\
  \\
      \textsuperscript{}Tsinghua University \quad
  }

\begin{document}

\maketitle
\ificlrfinal
\renewcommand{\thefootnote}{\fnsymbol{footnote}}
    \footnotetext[1]{Equal Contribution.}
    \footnotetext[2]{Work  done while YL, ZH, and YJ interned at Z.AI.}
\renewcommand{\thefootnote}{{$^\dagger$ footnote}}
\fi

\begin{abstract}
Long-horizon agentic LLMs are increasingly limited by finite context windows, as extended interaction trajectories can exceed the maximum context length before a task is completed. 
Context compaction offers a natural solution by summarizing previous interaction states and continuing the rollout under a compressed context, but incorporating compaction into reinforcement learning remains underexplored.
We propose \model{}, a reinforcement learning strategy to train long-horizon  agentic LLMs with context compaction. 
Our approach jointly optimizes \textit{task execution and summary generation} with token-level loss normalization and cross-segment generalized advantage estimation. 
This design enables the LLM agents to learn from compacted long-horizon trajectories. 
We train \model{} on top of open models and observe consistent performance gains on agentic coding tasks. 
\model{} enables the open GLM-4.5-Air model (106B-A12B) to achieve Pass@1 scores of  66.4\%  on SWE-bench Verified and  26.2\% on Terminal-Bench 2.0, exceeding the base model under  inference-time compaction by 6.6 and 4.9 points, respectively.
Built upon GLM-4.7-Flash (30B-A3B), \model{} improves Pass@1 by 5.5 and 6.7 points against the base model, reaching 56.0\% on SWE-bench Verified and 20.2\% on Terminal-Bench 2.0. 
\model{} is thus deployed in the RL pipeline for training the open GLM-5.2 model (750B-A40B). 

\end{abstract}

\section{Introduction}

Large language model (LLM) agents are increasingly applied to long-horizon interactive tasks, such as software engineering, terminal-based problem solving, and web interaction~\citep{yao2023react,nakano2021webgpt,jimenez2023swe}. These tasks require agents to repeatedly reason, act, observe environment feedback, and revise their plans over many steps. As the interaction proceeds, the accumulated history, including tool outputs, intermediate reasoning, error messages, and partial solutions, can exceed the model's finite context window. Although longer-context models alleviate this issue, scaling context length alone is costly and does not fully solve degraded utilization over long sequences~\citep{beltagy2020longformer,dao2022flashattention,liu2024lost,hsieh2024ruler}. Thus, long-horizon RL for LLM agents requires mechanisms that allow training to proceed under a fixed context budget.

Context compaction provides a natural way to address this limitation. When the history approaches the context limit, earlier interaction history can be summarized into a shorter state, after which the agent resumes from the compacted summary and a small amount of recent context. Related ideas have been explored in language-agent memory, reflection, and long-horizon interaction systems~\citep{shinn2023reflexion,park2023generative,sun2025contextfolding}. However, prior uses of compaction are mostly treated as inference-time heuristics or external memory operations. In RL training, compaction has a more fundamental role: once a summary replaces the original history, it determines what information is available for all subsequent actions. Therefore, task success depends not only on the execution policy, but also on the quality of the compaction policy.

Context compaction introduces additional challenges for asynchronous agentic RL. First, differences in compaction frequency and subsequent execution length increase variation in rollout generation time. Group-relative methods such as GRPO must wait for the slowest rollout in each group to finish before computing advantages, increasing policy lag and off-policy effects~\citep{hou2026singlerolloutasynchronousoptimizationagentic}. Second, jointly optimizing task execution and summarization requires including both execution and summary segments in training despite summary segments being substantially shorter, creating a severe length imbalance that can lead to disproportionate weighting under segment-level loss. Third, computing GAE independently within each segment places the shared terminal reward at local segment boundaries, artificially shortening its temporal distance from earlier actions and summaries and distorting temporal credit assignment.

In this work, we propose \model{}, a PPO-based reinforcement learning framework for building long-horizon agentic LLMs with trainable context compaction, as shown in Figure~\ref{fig:head}. 
\model{} incorporates compaction into rollout sampling, reconstructing the agent context from a summary once context budget is exhausted. 
Its core strategy is to jointly optimize task execution actions and summarization actions under the same final task reward. 
This enables RL training horizon to extend beyond the peak context length while keeping the actual working context budget fixed. Our contributions are summarized as follows:

\begin{figure}[t]
    \centering
    \begin{minipage}[t]{0.53\linewidth}
        \centering
        \includegraphics[width=\linewidth]{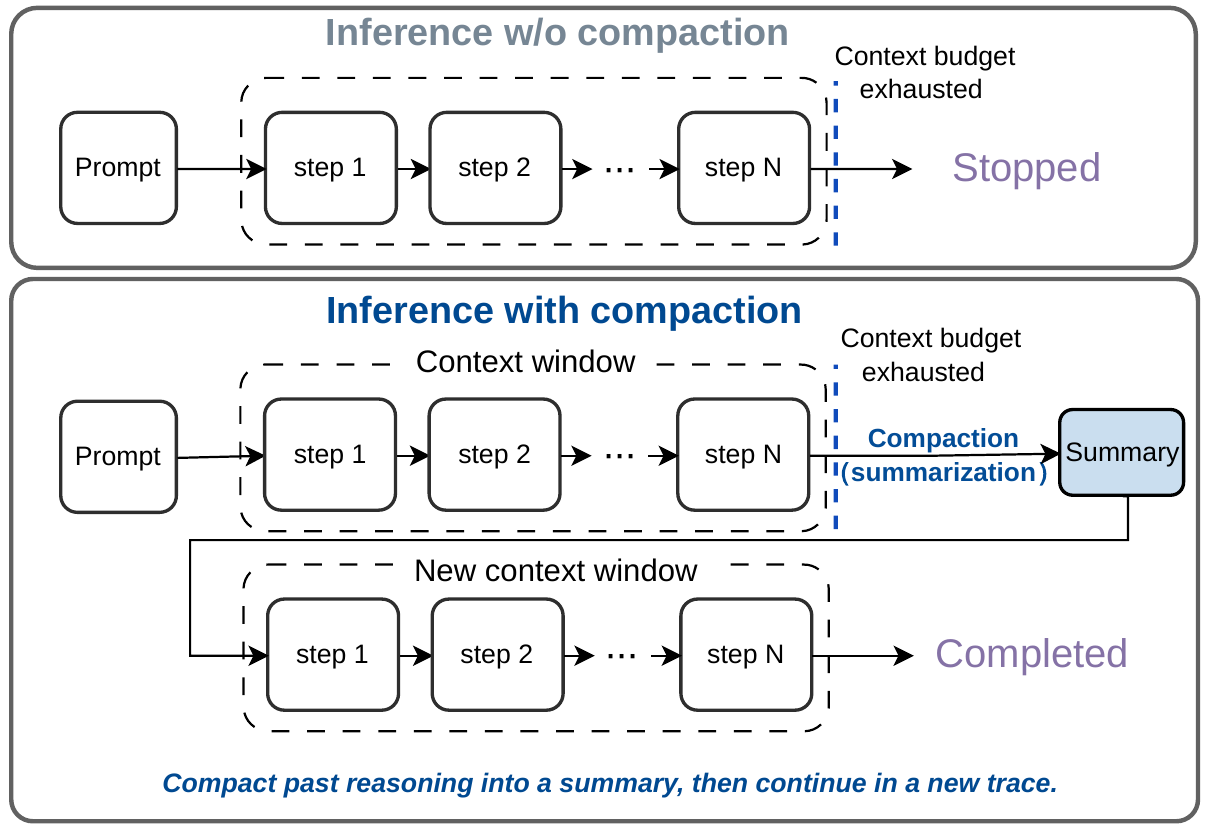}
    \end{minipage}
    \begin{minipage}[t]{0.42\linewidth}
        \centering
        \includegraphics[width=\linewidth]{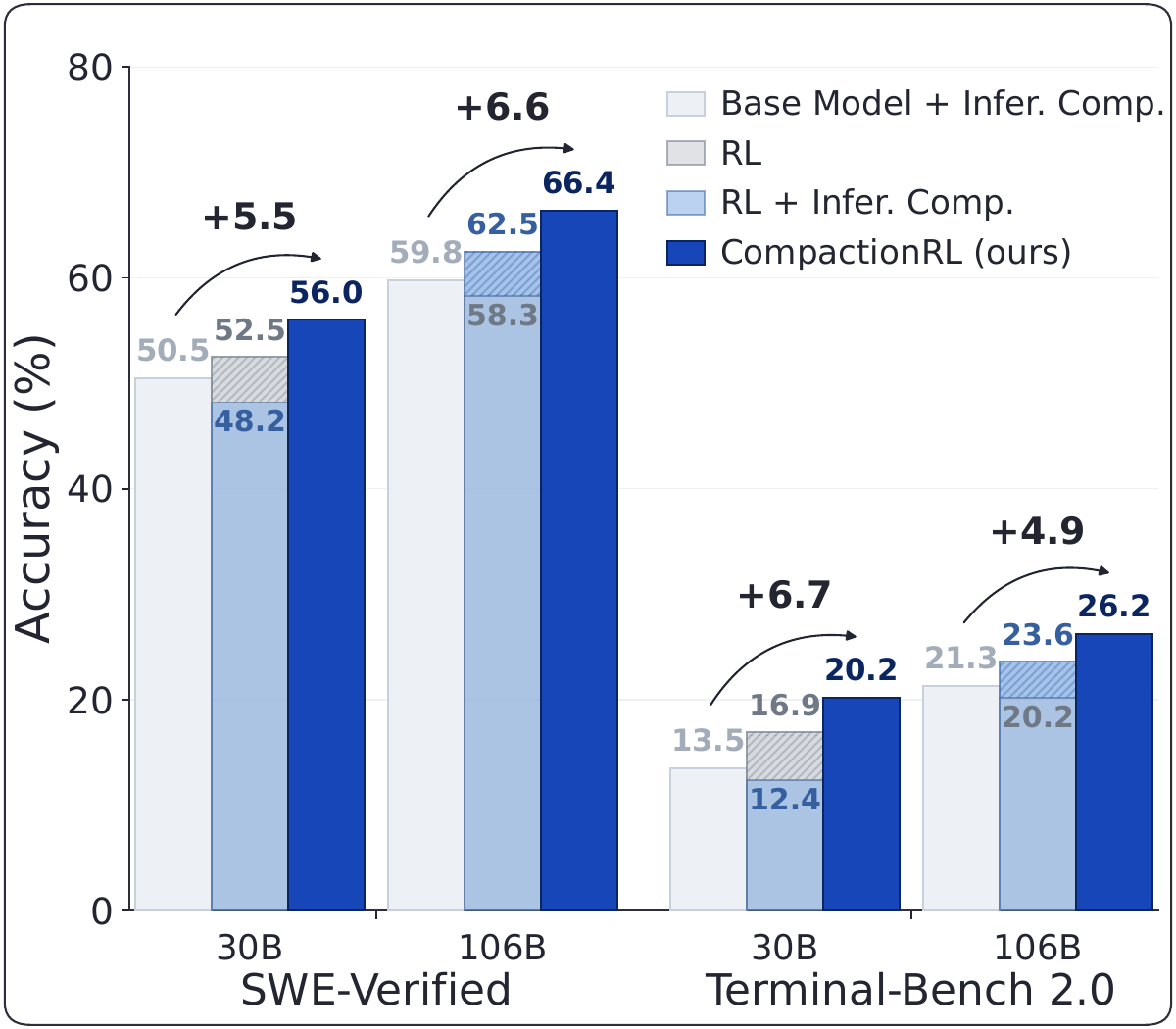}
    \end{minipage}
    \caption{
    Left: Context compaction allows execution to continue under a fixed context window by initiating a new trace from a compressed summary once the context budget is exhausted. Without compaction, execution terminates upon exhaustion of the context budget.
    Right: Performance on SWE-bench Verified and Terminal-Bench 2.0 across the base model with inference compaction, standard no-compaction RL, RL with inference compaction, and \model{}. ``Base Model + Infer. Comp.'' denotes the base model evaluated with context compaction. ``RL'' denotes standard RL trained and evaluated without compaction, while ``RL + Infer. Comp.'' uses the same RL checkpoint but enables context compaction during evaluation.
    }
    \label{fig:head}
\end{figure}

\begin{itemize}
    \item We bring context compaction into long-horizon agentic RL training. We adopt single-rollout PPO to avoid waiting for rollout groups, which is a problem more severe in compaction-aware training, and reduce the associated policy lag in asynchronous training, while using a learned critic to estimate token-level advantages for execution and summarization segments.

    \item We develop \model{} to jointly train trajectory generation and summary generation with a shared task-level reward, making summarization a learned part of the model rather than an inference-time heuristic. 
    To handle compacted rollouts, we use token-level loss normalization to reduce length-induced weighting bias and cross-segment GAE to preserve temporal credit assignment across compaction boundaries.

    \item \model{} achieves consistent gains on long-horizon agentic coding tasks. Compared with the base model with inference-time compaction, \model{} improves GLM-4.7-Flash by $5.5$ points on SWE-bench Verified and $6.7$ points on Terminal-Bench 2.0. 
    It also improves GLM-4.5-Air  by $6.6$ and $4.9$ points on the two coding benchmarks, respectively.
    These results show that RL-trained compaction can extend the effective training horizon of agentic LLMs without increasing the maximum working context length. 
\end{itemize}

\section{Related Work}

\paragraph{Context Compaction for Long-Horizon Agents.}
Long-horizon language agents must maintain task-relevant state across extended sequences of reasoning, tool calls, and environment feedback. A standard ReAct-style agent appends the full interaction history to the prompt~\citep{yao2023react}, but this makes the context grow with the trajectory and can reduce both efficiency and effective information use in long contexts~\citep{liu2024lost}. Prior work has addressed this issue through longer-context modeling and evaluation~\citep{bai2024longbench,hsieh2024ruler}, prompt or observation compression~\citep{jiang2023llmlingua,jiang2023longllmlingua}, and explicit memory or reflection mechanisms~\citep{packer2023memgpt,shinn2023reflexion,park2023generative, zhou2025mem1learningsynergizememory}, showing that compressing or externalizing history can extend the effective horizon of LLM agents.
Recent work has begun to incorporate context management into agent training. SUPO ~\citep{lu2025supo} and Memex(RL)~\citep{wang2026memexrlscalinglonghorizonllm} introduce context management into multi-turn RL and optimize with GRPO, without considering context state differences across compaction segments for advantage calculation; ReSum~\citep{wu2025resum} trains on segmented trajectories
without optimizing the summarizer while \cite{yi2026learningagentcompatiblecontextmanagement} and \cite{wu2026remembermattersproactivememory} train only the context manager.
However, how to jointly train long-horizon coding agents with both execution and compaction segments remains underexplored.

\paragraph{Reinforcement Learning for Language Models.}
Reinforcement learning has become a central technique for improving language models with outcome feedback\citep{uesato2022solvingmathwordproblems}. Early RLHF systems commonly adopt PPO with an explicit value function~\citep{schulman2017ppo,ouyang2022instructgpt}, while recent reasoning-oriented LLM training often favors critic-free or group-relative objectives such as RLOO, GRPO, and DAPO~\citep{ahmadian2024back,shao2024deepseekmath,deepseekai2025deepseekr1,hou2025t1advancinglanguagemodel, yu2025dapo, zheng2025group} that have shown strong performance on verifiable-reward tasks such as mathematical reasoning and coding\citep{hou2025treerlllmreinforcementlearning, wei2025swerladvancingllmreasoning, xu2025mobilerlonlineagenticreinforcement}.
At the same time, recent studies show that critic-based methods can still be highly effective when value learning is carefully designed. VC-PPO~\citep{yuan2025vcppo} improves PPO through value pretraining and modified advantage estimation and VAPO~\citep{yue2025vapo} demonstrates that a value-based RL framework can achieve strong and stable reasoning performance. SAO~\citep{hou2026singlerolloutasynchronousoptimizationagentic} adopts single-rollout sampling to reduce policy lag caused by waiting for complete rollout groups in asynchronous agentic RL. These studies motivate critic-based optimization for long-horizon agents, supporting both state-dependent advantage estimation and training without group-relative baselines.

\section{The \model{} Method}
\label{sec:method}

\begin{figure}
    \centering
    \includegraphics[width=\linewidth]{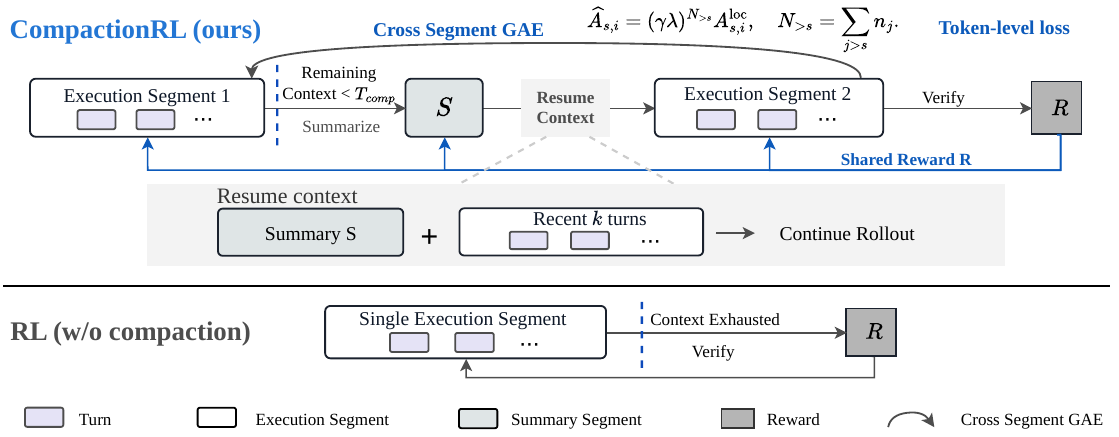}
    \caption{Overview of \model{}. During rollout collection, the agent interacts with the environment under a fixed context budget. When the remaining context budget falls below a compaction threshold $T_{\mathrm{comp}}$, the policy generates a summary and resumes from the summary and recent turns. All execution and summary segments are optimized and share the trajectory reward $R$. During training, we propagate this reward across segment boundaries by correcting the local GAE advantage $A^{\mathrm{loc}}_{s,i}$ with the number of optimized tokens in subsequent segments, $N_{>s}$, yielding the cross-segment advantage $\widehat{A}_{s,i}$. Standard RL without compaction optimizes each rollout as a single execution segment.
    }
    \label{fig:overview}
\end{figure}

We present \model{}, a reinforcement learning framework for long-horizon language agents under a fixed context budget. The central idea is to make context compaction part of rollout sampling: when the interaction history approaches the context limit, the agent produces a compact summary of the previous trajectory and then resumes from a reconstructed context containing the summary and a short history tail of recent interaction. The summary is generated by the trainable policy and is optimized together with ordinary task-execution actions under the final task reward. Figure~\ref{fig:overview} gives an overview of the framework.

\subsection{Trainable Context Compaction}
\label{sec:trainable_compaction}

\paragraph{Compaction during Rollout Sampling.}
We represent the interaction history as
\begin{equation}
    h_t = (s, u, z_1, \ldots, z_t),
    \qquad
    z_i = (a_i, o_i),
\end{equation}
where $s$ is the system prompt, $u$ is the original user instruction, $a_i$ is the assistant response at step $i$, and $o_i$ is the corresponding environment observation. We treat each assistant-observation pair $z_i$ as an atomic step, so that tool calls and their feedback are not separated by compaction.

Let $C$ denote the context budget and $|h_t|$ the token length of the current history, Compaction is triggered when the remaining context budget falls below a threshold $T_{\mathrm{comp}}$:
\begin{equation}
    C - |h_t| < T_{\mathrm{comp}}.
\end{equation}
Once triggered, we append a fixed summarization instruction $q_{\mathrm{sum}}$ to the current history and sample a summary $S_t$ from the policy,
where $q_{\mathrm{sum}}$ asks the model to preserve information necessary for continuing the task, including the original goal, completed actions, important observations, unresolved errors, current state, and plausible next steps.

After summary generation, the rollout continues from a reconstructed context
\begin{equation}
    \bar{h}_t
    =
    s
    \oplus
    u_{\mathrm{resume}}(S_t)
    \oplus
    (z_{t-k+1}, \ldots, z_t),
\end{equation}
where $u_{\mathrm{resume}}(S_t)$ is a fixed template containing the generated summary. The final term keeps the most recent $k$ turns while the summary covers only the history preceding the $k$ turns. We use $k=2$ by default and reduce $k$ when necessary to ensure that $\bar{h}_t$ fits within the context budget. This reconstruction preserves long-range information through $S_t$ while retaining the most recent environment state exactly.

\paragraph{Training Segments.}
Compaction naturally partitions a rollout into a sequence of trainable segments,
\begin{equation}
    \tau = (\sigma_1, \ldots, \sigma_K),
\end{equation}
where each segment is either an execution segment or a summarization segment. An execution segment contains tokens produced for solving the task, while a summarization segment contains the summary tokens generated before context reconstruction. Summary tokens are sampled from the same trainable policy and are included in the RL objective.

Each rollout receives a final task reward $R(\tau)$ determined by task correctness. We assign this rollout-level reward to all trainable segments from the same rollout. We do not introduce a separate summary-quality reward, since hand-designed summary metrics may not reflect which details are useful for continuation.


\paragraph{Effect of Summarization Quality.}
Compaction makes task success sensitive to the quality of the summary. A summary that omits a crucial file path, error message, failed command, or partial solution can make subsequent actions ineffective even when the execution policy itself is strong. To quantify this effect, we fix the execution agent and vary only the summary agent.

\begin{table}[t]
    \centering
    \caption{Effect of summarization quality under context compaction. The execution agent is fixed to GLM-4.7-Flash, while tested with various summary agents including Qwen3.5-27B\citep{qwen3.5} and Qwen3-30B-A3B\citep{yang2025qwen3}. SWE-Verified Acc. denotes pass@1(\%) on SWE-bench Verified; Summary Count / Trace reports the average number of summaries triggered per trace.}
    \label{tab:summary_agent_ablation}
    \setlength{\tabcolsep}{6pt}
    \begin{tabular}{@{}cccc@{}}
        \toprule
        Execution Agent & Summary Agent & SWE-Verified Acc. & Summary Count / Trace \\
        \midrule
        GLM-4.7-Flash & Qwen3.5-27B      & 55.5 & 1.010 \\
        GLM-4.7-Flash & GLM-4.7-Flash    & 50.5 & 1.075 \\
        GLM-4.7-Flash & Qwen3-30B-A3B     & 49.0 & 1.126 \\
        \bottomrule
    \end{tabular}
\end{table}

Table~\ref{tab:summary_agent_ablation} shows that changing only the summary agent leads to a large difference in final task performance. The best summarizer improves SWE-Verified accuracy from $49.0$ to $55.5$, a gain of $6.5$ absolute points, while also triggering slightly fewer compactions per trace. This result suggests that compaction is a performance-critical decision process rather than a passive preprocessing step. We therefore train summary generation jointly with task execution.

\subsection{Optimization under Compacted Rollouts}

\paragraph{Single-Rollout Optimization.}
Compaction increases variation in rollout generation time since rollouts differ in compaction frequency and subsequent execution length. Group-relative advantage estimation requires completed rollouts to wait for the slowest member of the group, while asynchronous policy updates increase the mismatch between rollout and training policies during this waiting period. Following SAO~\citep{hou2026singlerolloutasynchronousoptimizationagentic}, we adopt PPO with one rollout per prompt to avoid this additional waiting and reduce the associated off-policy effects.

Moreover, resumed segments condition on different compacted histories. A learned critic provides context-dependent value estimates for token-level advantage estimation, whereas rollout-level group normalization does not explicitly account for these intermediate states differences.

\paragraph{Token-Level Loss for Length Imbalance.}
Since the length of agent trajectories and summarization varies significantly, we choose to use token-level loss instead of sequence-level loss to overcome the severe imbalance.
Let $\mathcal{M}$ denote the set of optimized assistant-token positions in a training batch. For token $y_{s,i}$ in segment $\sigma_s$, conditioned on context $x_{s,i}$, let $\rho_{s,i}(\theta)$ denote the importance sampling ratio of the current policy $\pi_\theta$ to the rollout policy $\pi_{\theta_{\mathrm{old}}}$. We optimize the clipped PPO objective
\begin{equation}
    \mathcal{L}_{\pi}
    =
    -\frac{1}{|\mathcal{M}|}
    \sum_{(s,i)\in\mathcal{M}}
    \min
    \left(
        \rho_{s,i}(\theta)\widehat{A}_{s,i},
        \mathrm{clip}
        \left(
            \rho_{s,i}(\theta),
            1-\epsilon,
            1+\epsilon
        \right)
        \widehat{A}_{s,i}
    \right),
\end{equation}
where $\epsilon$ is the clipping threshold and $\widehat{A}_{s,i}$ is the advantage estimate described below.

The loss computation is performed over generated tokens rather than over samples. This is important because compaction introduces heterogeneity in both segment count and segment length. Segment-level averaging would make a rollout with more compaction events contribute more strongly than a rollout with fewer segments, even when both receive the same task reward. Token-level normalization removes this segment-count bias and gives each trainable token equal weight.

\paragraph{Cross-Segment Generalized Advantage Estimation.}
PPO~\citep{schulman2017ppo} uses a learned critic $V_\phi$ to estimate expected discounted returns. The temporal-difference (TD) residual measures the discrepancy between the one-step bootstrapped return and the current value estimate:
\begin{equation}
    \delta_t
    =
    r_t + \gamma V_\phi(x_{t+1}) - V_\phi(x_t),
\end{equation}
where $x_t$ is the current context, $r_t$ is the reward, and $\gamma$ is the discount factor. Generalized advantage estimation (GAE)~\citep{schulman2015high} computes token-level advantages by accumulating subsequent TD residuals:
\begin{equation}
    A_t^{\mathrm{GAE}}
    =
    \sum_{\ell=0}^{T-t}
    (\gamma\lambda)^\ell\delta_{t+\ell},
\end{equation}
where $T$ is the number of optimized tokens and $\lambda$ is the trace parameter. This accumulation propagates delayed task feedback to earlier tokens while discounting more distant residuals.

Compacted rollouts also create a temporal credit-assignment issue. Since each segment is optimized independently, a naive segment-level GAE computation places the shared terminal reward at the end of every segment. For earlier segments, this makes the final task outcome appear artificially closer than it is in the original rollout, over-crediting actions or summaries that occur far before task completion.

For a segment $\sigma_s$ with $n_s$ optimized tokens, the local GAE estimator is
\begin{equation}
    A^{\mathrm{loc}}_{s,i}
    =
    \sum_{\ell=0}^{n_s-i}
    (\gamma\lambda)^\ell
    \delta_{s,i+\ell},
    \qquad
    \delta_{s,i}
    =
    r_{s,i}
    +
    \gamma V_\phi(x_{s,i+1})
    -
    V_\phi(x_{s,i}),
\end{equation}
where $i$ indexes tokens within the segment. Let
    $N_{>s} = \sum_{j>s} n_j$
be the number of optimized tokens generated after segment $\sigma_s$ in the same rollout. We apply a trajectory-position correction
\begin{equation}
    \widehat{A}_{s,i}
    =
    (\gamma\lambda)^{N_{>s}}
    A^{\mathrm{loc}}_{s,i}.
\end{equation}
This correction discounts earlier segments according to the number of subsequent trainable tokens before rollout termination. In particular, if the terminal task reward appears at the last token of each independently optimized segment, then the reward term for token $(s,i)$ receives discount
\begin{equation}
    (\gamma\lambda)^{N_{>s}+n_s-i},
\end{equation}
which matches its distance to the final outcome in the concatenated compacted rollout.

Note that this is an approximation rather than an exact reconstruction of full-trajectory GAE. Since an exact reconstruction of full-trajectory GAE would require recursively propagating TD residuals from later to earlier compaction segments, introducing cross-segment dependencies that preclude parallel processing within a batch, we use the trajectory-position correction above instead of exact reconstruction to preserve training efficiency.

\section{Experiments}

\subsection{Experimental Setup}

\textbf{Training Details.} 
We conduct experiments with two models at different parameter scales: GLM-4.7-Flash and GLM-4.5-Air-SFT. The latter is obtained by supervised fine-tuning GLM-4.5-Air \citep{5team2025glm45agenticreasoningcoding} on trajectories generated by GLM-4.7. For each model, we initialize the corresponding critic from the same model checkpoint and perform 50 steps of value pretraining on the training dataset before reinforcement learning. We use open-source training data from SWE-Dev \citep{wang2025swe}. For the training of \model{}, we use slime \citep{slime_github}, an open-source asynchronous RL framework, for reinforcement learning training.

For RL training, we employ a global batch size of 128, sample one rollout per prompt, and use context budgets of 64k for GLM-4.7-Flash and 80k for GLM-4.5-Air-SFT. The policy is optimized with Adam using a learning rate of $2\times 10^{-6}$, while the critic is trained with a learning rate of $3\times 10^{-6}$. For advantage estimation, we set the reward discount factor $\gamma=1$ and adopt length-adaptive GAE~\citep{yue2025vapo} with $\lambda = 1 - \frac{1}{\alpha l}$ and $\alpha=1.5$, where $l$ denotes the response length. To ensure that the critic keeps pace with policy updates, we perform two value model updates and one policy update for each batch, facilitating accurate advantage estimation. Each assistant response is limited to 10,240 tokens and compaction is triggered when the remaining context budget falls below 10,240 tokens, with at most three compaction operations per rollout.
Detailed model configurations and compute resources are provided in
Appendix~\ref{app:training_compute_details}.

\textbf{Evaluation Setting.} 
We evaluate \model{} on SWE-bench Verified \citep{jimenez2023swe} and Terminal-Bench 2.0 \citep{merrill2026terminalbenchbenchmarkingagentshard}, using 200 randomly sampled tasks from SWE-bench Verified and the full Terminal-Bench 2.0 set, and report Pass@1 accuracy. The complete list of sampled tasks is provided in
Appendix~\ref{app:swe_verified_subset}. All evaluations are conducted in the Harbor \citep{Harbor_Framework} environment with the Terminus-KIRA \citep{terminuskira2026} agent scaffold. Evaluation hyperparameters include: top-$p=1.0$, temperature $=1.0$, up to 250 interaction turns and at most three compaction operations per trajectory.

\subsection{Main Results}

Table~\ref{tab:main_results_swe_tb} compares the base models, standard no-compaction          PPO , \model{} (w/o sum. loss), and the full \model{} under single-window or compacted evaluation settings on SWE-bench Verified and Terminal-Bench 2.0. \model{} (w/o sum. loss) enables compaction during RL but masks the loss on summary-response turns, separating the effect of training on compacted histories from that of directly optimizing the summarizer.

\model{} achieves the best performance across both model scales and benchmarks. Inference-time compaction generally improves the base models by extending their effective context budget. Standard PPO improves single-window execution, but this benefit does not transfer consistently to compacted inference, where it either degrades performance or yields smaller gains than \model{}. Moreover, the consistent improvements of \model{} over \model{} (w/o sum. loss) across both model scales and benchmarks demonstrate the benefit of directly optimizing summarization during compaction-aware training. Overall, these results indicate that \model{} effectively extends the usable context horizon through trainable compaction.

\begin{table}[t]
    \centering
    \caption{
    Main results on SWE-bench Verified and Terminal-Bench 2.0.
    We report Pass@1 accuracy (\%). Results are reported as mean $\pm$ standard deviation over three evaluation runs. Comp. Training and Comp. Infer. indicate whether context compaction is enabled during RL training and evaluation, respectively. For each model, ``w/ Inference Compaction'' uses the same checkpoint as the preceding row but enables compaction during evaluation. \model{} (w/o sum. loss) retains compaction during both training and inference but masks summary-response tokens in the RL loss. Compacted evaluation retains the peak working context length and allows up to three compactions, yielding an effective budget of up to $4\times$ the peak length.
    }
    \label{tab:main_results_swe_tb}
    \setlength{\tabcolsep}{3.6pt}
    \renewcommand{\arraystretch}{1.05}
    \newcommand{\std}[1]{{\scriptstyle \pm #1}}
    {
    \begin{tabular}{@{}lc cc cc@{}}
        \toprule
         & Peak Len. & \makecell{Comp.\\Training} & \makecell{Comp.\\Infer.} & \makecell{SWE-bench \\Verified} & \makecell{Terminal-Bench 2.0} \\
        \midrule
        \multicolumn{6}{l}{\textit{Public reports / baselines}} \\
        \midrule
        gpt-oss-120b
            & 128k & -- & \xmark
            & 62.0 & 18.7 \\
        Qwen3.5-35B-A3B
            & 64k & -- & \cmark
            & 58.0 & 27.0 \\
        Qwen3-Coder-480B-A35B-Instruct
            & 256k & -- & \xmark
            & 66.5 & 23.9 \\
        Qwen3-Coder-30B-A3B-Instruct
            & 256k & -- & \xmark
            & 51.9 & 14.6 \\
        Qwen3-235B-A22B-Instruct-2507
            & 256k & -- & \xmark
            & 45.2 & 13.5 \\
        Qwen3-30B-A3B-Instruct-2507
            & 256k & -- & \xmark
            & 25.2 & 5.34 \\
        \midrule
        \multicolumn{6}{l}{\textit{Terminus-KIRA scaffold}} \\
        \midrule
        GLM-4.7-Flash (30B-A3B)
            & 64k & -- & \xmark
            & $47.3\std{2.7}$ & $14.6\std{1.6}$ \\
        \quad \textcolor{gray}{w/ Inference Compaction}
            & 64k & -- & \cmark
            & $50.5\std{1.2}$ & $13.5\std{0.9}$ \\
        + RL (w/o compaction)
            & 64k & \xmark & \xmark
            & $52.5\std{2.5}$ & $16.9\std{2.4}$ \\
        \quad \textcolor{gray}{w/ Inference Compaction}
            & 64k & \xmark & \cmark
            & $48.2\std{0.5}$ & $12.4\std{0.9}$ \\
        + \model{} (w/o sum. loss)
            & 64k & \cmark & \cmark
            & $54.5\std{1.2}$ & $12.4\std{0.0}$ \\
        \rowcolor{blue!8}
        + \model{} (ours)
            & 64k & \cmark & \cmark
            & \textbf{\bm{$56.0\std{0.4}$}} & \textbf{\bm{$20.2\std{0.9}$}} \\
        \midrule
        GLM-4.5-Air (106B-A12B)
            & 80k & -- & \xmark
            & $57.8\std{2.1}$ & $18.0\std{2.8}$ \\
        \quad \textcolor{gray}{w/ Inference Compaction}
            & 80k & -- & \cmark
            & $59.8\std{1.9}$ & $21.3\std{1.6}$ \\
        + RL (w/o compaction)
            & 80k & \xmark & \xmark
            & $58.3\std{0.6}$ & $20.2\std{3.7}$ \\
        \quad \textcolor{gray}{w/ Inference Compaction}
            & 80k & \xmark & \cmark
            & $62.5\std{1.1}$ & $23.6\std{0.9}$ \\
        + \model{} (w/o sum. loss)
            & 80k & \cmark & \cmark
            & $64.5\std{2.4}$ & $21.7\std{2.8}$ \\
        \rowcolor{blue!8}
        + \model{} (ours)
            & 80k & \cmark & \cmark
            & \textbf{\bm{$66.4\std{0.7}$}} & \textbf{\bm{$26.2\std{2.1}$}} \\
        \bottomrule
    \end{tabular}
    }

    \vspace{2pt}
    \begin{minipage}{0.96\linewidth}
    \footnotesize
    \textit{Note.} In all our results, SWE-bench Verified is evaluated on a random 200-instance subset, while public baselines are reported on the full benchmark. The public baselines are taken from model cards or public reports and are included for reference only, as their agent scaffolds may differ from ours.
    \end{minipage}
\end{table}

\subsection{Ablation Studies}
\paragraph{Effect of PPO.}
To evaluate the role of PPO in compaction-aware training, we compare \model{} with a GRPO-based baseline implemented following SUPO~\citep{lu2025supo}.
Both methods jointly optimize task execution and summary generation, but SUPO computes group-relative advantages at the rollout level and assigns them uniformly to all compaction segments, whereas \model{} uses a learned critic to estimate token-level advantages.
As shown in Table~\ref{tab:optimization_ablation}, \model{} outperforms the SUPO baseline by 4.6 and 4.9 percentage points on SWE-bench Verified and Terminal-Bench 2.0, respectively, demonstrating the effectiveness of critic-based PPO training.

\paragraph{Effect of Token-Level Loss and Cross-Segment GAE.}
We further ablate the two optimization components designed for compacted trajectories using GLM-4.5-Air-SFT: token-level loss normalization and cross-segment GAE. 
As shown in Table~\ref{tab:optimization_ablation}, removing either component degrades performance, confirming that both loss weighting and cross-segment credit assignment are important for compaction-aware RL. A larger degradation is observed when token-level loss normalization is removed, suggesting that correcting the optimization bias induced by variable segment counts and lengths is particularly critical. 

\begin{table}[t]
    \centering
    \caption{Ablation study on PPO based RL, token-level loss normalization, and cross-segment GAE using GLM-4.5-Air-SFT. All variants are evaluated under the compaction enabled 80k$\times$4 setting. We report Pass@1 accuracy as mean ± standard deviation on SWE-bench Verified and Terminal-Bench 2.0.}
    \label{tab:optimization_ablation}
    \setlength{\tabcolsep}{8pt}
    \newcommand{\std}[1]{{\scriptstyle \pm #1}}
    \begin{tabular}{@{}lcc@{}}
        \toprule
         & SWE-bench Verified & Terminal-Bench 2.0 \\
        \midrule
        GLM-4.5-Air (106B-A12B) & $59.8\std{1.9}$ & $21.3\std{1.6}$ \\
        + SUPO & $61.8\std{1.8}$ & $21.3\std{0.9}$ \\
        \rowcolor{blue!8}
        + \model{} & \textbf{\bm{$66.4\std{0.7}$}} & \textbf{\bm{$26.2\std{2.1}$}} \\
        \quad $-$ w/o token-level loss & $60.0\std{1.2}$ & $21.3\std{2.4}$ \\
        \quad $-$ w/o cross-segment GAE & $63.0\std{1.4}$ & $22.5\std{2.4}$ \\
        \bottomrule
    \end{tabular}
\end{table}

\paragraph{Effect of Extended-Context.}
We further compare \model{} with standard no-compaction RL using an extended context window of 160k, which simulates access to a sufficiently long context that can retain the full trajectory without compaction-induced information loss. As shown in Table~\ref{tab:extended_context_training}, under 80k$\times$4 compacted evaluation, \model{} outperforms the extended-context baseline evaluated with either 80k$\times$4 compaction or a 160k non-compacted context on both benchmarks. This demonstrates the effectiveness of trainable compaction relative to directly increasing the training context window.

\begin{table}[t]
    \centering
    \caption{Comparison with extended-context no-compaction RL using GLM-4.5-Air (106B-A12B). We report Pass@1 accuracy (\%) as mean $\pm$ standard deviation over three evaluation runs. As inference time strategies, Comp.($\times 4$) enables up to three compaction operations with an 80k peak context, while Long uses a 160k non-compacted context window as a sufficiently long context reference.}
    \label{tab:extended_context_training}
    \setlength{\tabcolsep}{3.6pt}
    \renewcommand{\arraystretch}{1.05}
    \newcommand{\std}[1]{{\scriptstyle \pm #1}}
    {
    \begin{tabular}{@{}lc cc cc@{}}
        \toprule
        \multirow{2}{*}{} &
        \multirow{2}{*}{\makecell{Train\\Budget}} &
        \multicolumn{2}{c}{SWE-bench Verified} &
        \multicolumn{2}{c}{Terminal-Bench 2.0} \\
        \cmidrule(lr){3-4} \cmidrule(lr){5-6}
        & &
        \makecell{Comp.($\times 4$)} & Long &
        \makecell{Comp.($\times 4$)} & Long \\
        \midrule
        GLM-4.5-Air (106B-A12B)
            & --
            & $59.8\std{1.9}$ & $59.5\std{1.6}$
            & $21.3\std{1.6}$ & $20.6\std{2.1}$ \\
        + RL (w/o compaction)-160k
            & 160k
            & $62.8\std{2.8}$ & $64.0\std{0.8}$
            & $23.2\std{2.8}$ & $22.1\std{1.4}$ \\
        \rowcolor{blue!8}
        + \model{} (ours)
            & 80k$\times$4
            & \textbf{\bm{$66.4\std{0.7}$}} & $61.8\std{1.5}$
            & \textbf{\bm{$26.2\std{2.1}$}} & $24.2\std{1.4}$ \\
        \bottomrule
    \end{tabular}
    }
\end{table}

\subsection{Analysis}
\paragraph{Compaction Behavior.}

\begin{figure}[t]
    \centering
    \begin{minipage}[t]{0.32\linewidth}
        \centering
        \includegraphics[width=\linewidth]{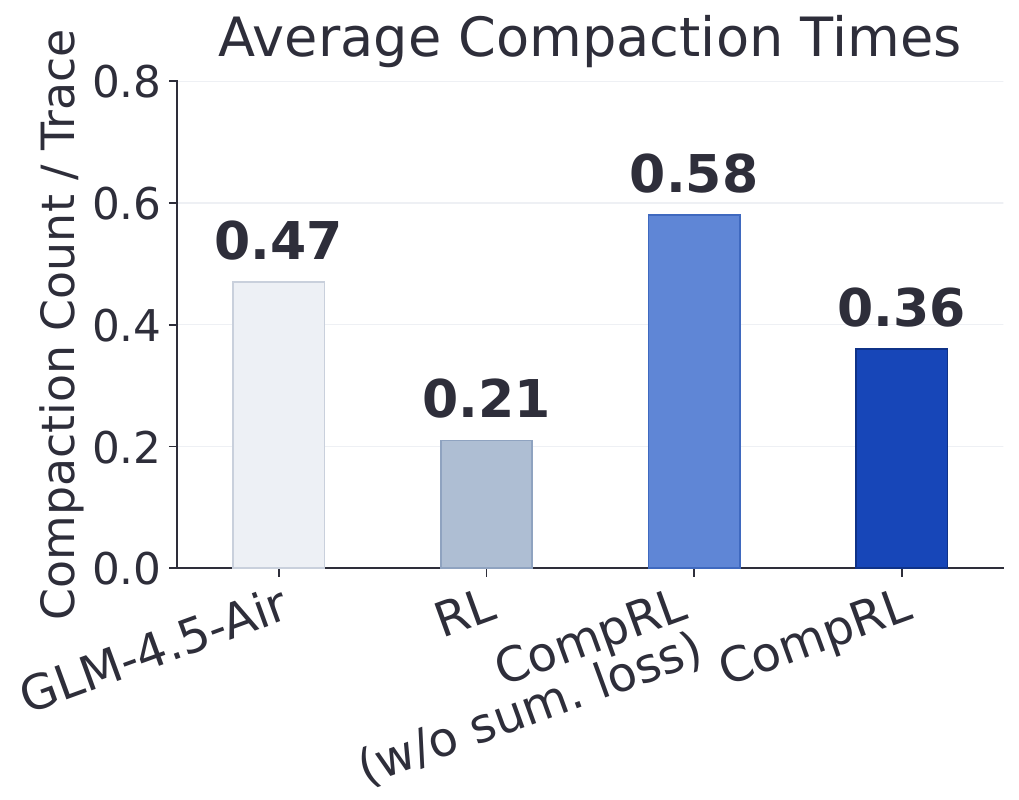}
        {\small (a) Average compactions}
    \end{minipage}%
    \hspace{0.01\linewidth}%
    \begin{minipage}[t]{0.32\linewidth}
        \centering
        \includegraphics[width=\linewidth]{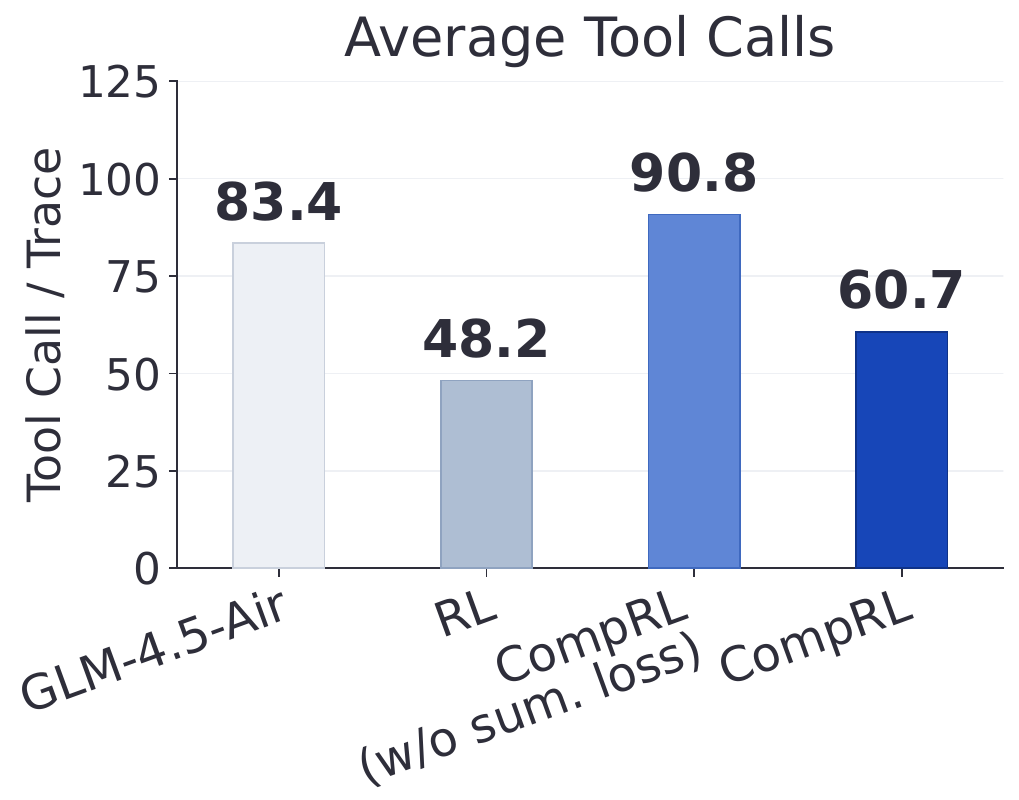}
        {\small (b) Average tool call counts}
    \end{minipage}%
    \hspace{0.01\linewidth}%
    \begin{minipage}[t]{0.32\linewidth}
        \centering
        \includegraphics[width=\linewidth]{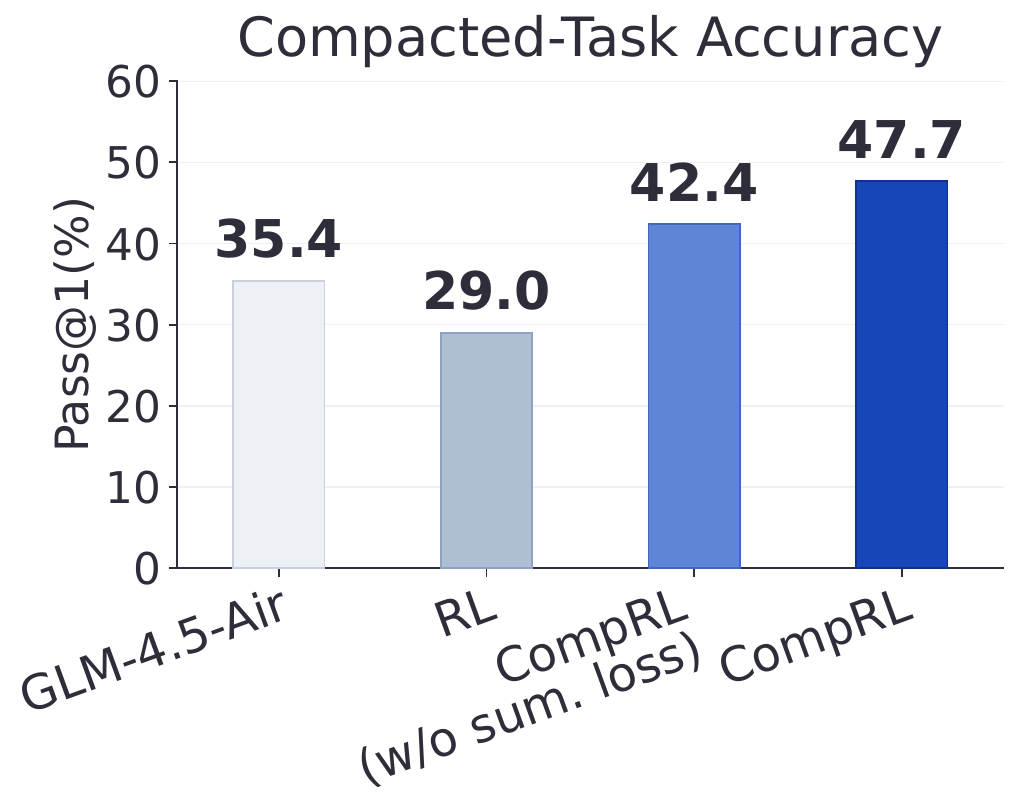}
        {\small (c) Compacted-task accuracy}
    \end{minipage}
    \caption{Behavior comparison on SWE-bench Verified in the GLM-4.5-Air setting under 80k$\times$4 compacted evaluation. We compare the base model(GLM-4.5-Air-SFT), standard no-compaction RL, \model{} (w/o sum. loss), and \model{}. Here, we denote a trace as all compaction segments generated to solve a single task. (a) Average number of compaction operations per trace. (b) Average number of tool calls per trace. (c) Pass@1 accuracy on compaction triggering tasks in each evaluation run.}
    \label{fig:compaction_behavior}
\end{figure}

Figure~\ref{fig:compaction_behavior} analyzes compaction behavior in the GLM-4.5-Air setting under 80k$\times$4 evaluation. Figure~\ref{fig:compaction_behavior}(a,b) shows that \model{} requires fewer compactions and tool calls than GLM-4.5-Air and \model{} (w/o sum. loss) but more than standard no-compaction RL. This indicates that \model{} learns both to exploit the extended context horizon enabled by compaction and a higher interaction efficiency, suggesting that the improvement is not attributable to longer interaction traces alone; rather, trained summaries help retain task-relevant information and reduce redundant re-exploration after compaction. Figure~\ref{fig:compaction_behavior}(c) reports pass@1 on tasks that trigger compaction. \model{} achieves the highest accuracy on this subset, indicating that jointly training execution and summarization improves reliable continuation from compacted histories.

\paragraph{Training Dynamics.}

\begin{figure}[t]
    \centering
    \begin{minipage}[t]{0.33\linewidth}
        \centering
        \includegraphics[width=\linewidth]{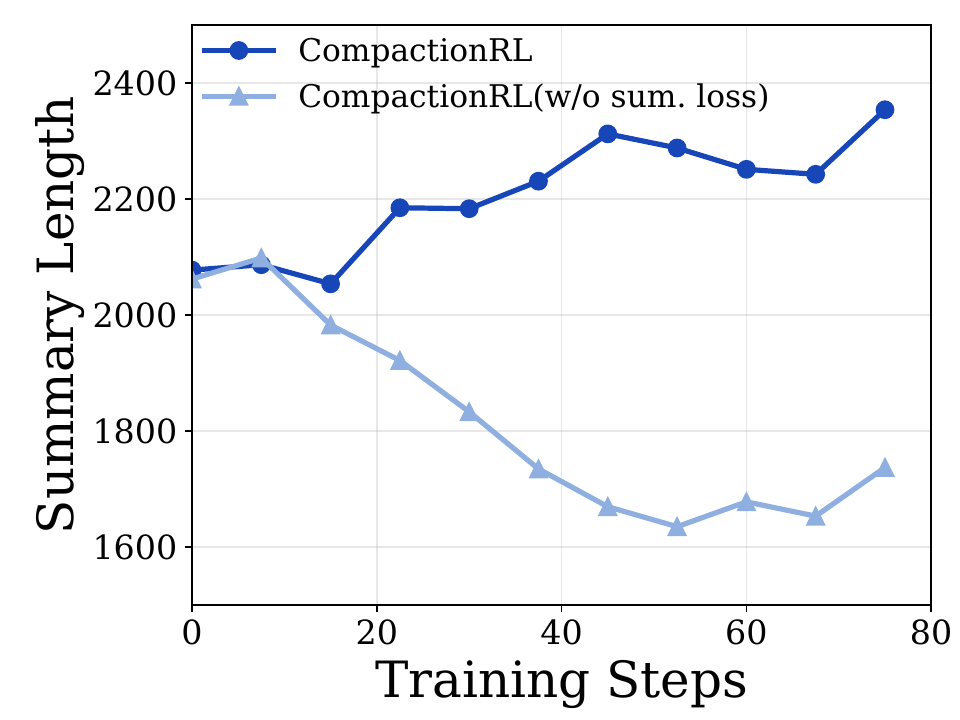}
        {\small (a) Summary length}
    \end{minipage}%
    \begin{minipage}[t]{0.33\linewidth}
        \centering
        \includegraphics[width=\linewidth]{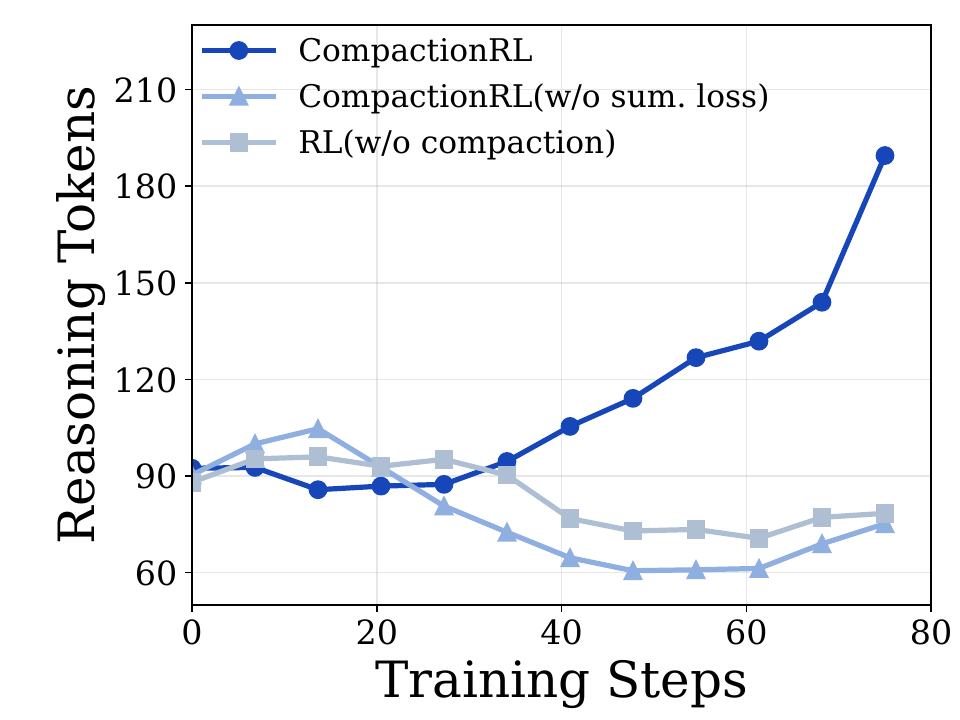}
        {\small (b) Reasoning tokens per turn}
    \end{minipage}%
    \begin{minipage}[t]{0.33\linewidth}
        \centering
        \includegraphics[width=\linewidth]{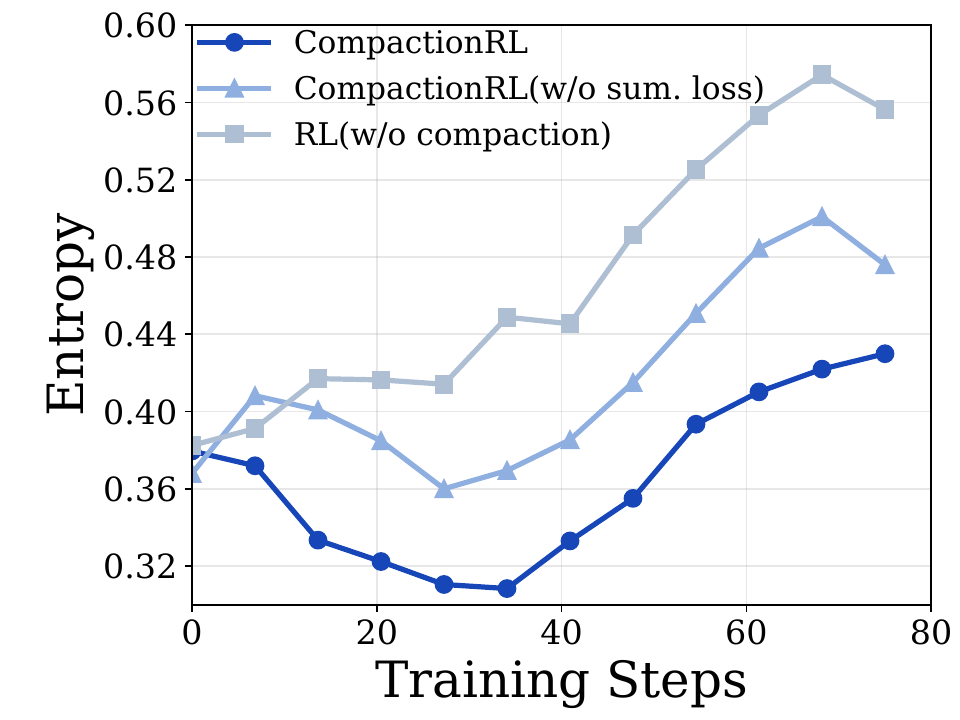}
        {\small (c) Entropy}
    \end{minipage}
    \caption{Training dynamics of GLM-4.5-Air \model{}. (a) Summary optimization leads to increasingly longer and more detailed summaries. (b) \model{} increases reasoning tokens per turn. (c) Policy entropy.}
    \label{fig:training_dynamics}
\end{figure}

Figure~\ref{fig:training_dynamics}(a) analyzes the change of summary length during GLM-4.5-Air \model{} training. Compared with \model{} (w/o sum. loss), whose summary length decreases during training, \model{} produces increasingly longer and more detailed summaries  when summary responses are included in the RL objective. This results in more complete and actionable compacted states that better preserve implementation-relevant context and continuation state. This suggests that optimizing summary responses encourages the model to retain useful detailed information for post-compaction continuation, rather than merely recording high-level task progress. We further quantify summary faithfulness and continuation usefulness in
Appendix~\ref{app:summary_quality_analysis}, supporting that \model{} learns to generate summaries of higher quality.  Figure~\ref{fig:training_dynamics}(b) further shows that the reasoning tokens of \model{} increase throughout training, whereas those of \model{} (w/o sum. loss) and standard no-compaction RL decrease. This indicates that learned compaction effectively expands the usable context window, allowing for more reasoning budget. Figure~\ref{fig:training_dynamics}(c) shows that \model{} also exhibits a slower entropy increase, suggesting more controlled policy optimization.

Additional analysis of hyperparameter sensitivity is provided in Appendix~\ref{app:compaction_sensitivity}.

\section{Conclusion}
We presented \model{}, a PPO-based framework that trains long-horizon agents with context compaction by optimizing execution and summary tokens under the same task reward. With token-level loss normalization and cross-segment GAE, \model{} improves SWE-bench Verified and Terminal-Bench 2.0 performance under a fixed peak context budget, while further ablations show that summary training produces more detailed compacted states and is important for these gains. Overall, our results demonstrate the effectiveness of incorporating context compaction into RL training for long-horizon agents, rather than treating it solely as an inference-time heuristic.
\paragraph{Limitations.}

\model {} is designed for settings where context compaction is available during both training and inference. Extending its benefits to single-window inference requires further study, as the gains observed under compaction-enabled evaluation do not consistently transfer to this setting. In addition, cross-segment GAE remains an approximation  to full-trajectory credit assignment, motivating further investigation into more precise reconstruction.

\subsection*{AI use statement}
We did not use generative AI tools for any tasks with required disclosure. We used generative AI tools only for language polishing and grammar correction. All research ideas, methods, experiments, and analyses were conceived, designed, implemented, and validated by the authors. We reviewed all AI-assisted text and revised it as necessary. We take responsibility for the final content of this work, including text produced with the aid of generative AI.

\bibliographystyle{iclr2027/iclr2027_conference}
\bibliography{references}

\appendix
\section{Experimental Details}
\label{app:experimental_details}

\subsection{SWE-bench Verified Subset}
\label{app:swe_verified_subset}

All SWE-bench Verified results in our experiments use the same
200-task subset spanning 10 repositories. For reproducibility,
Table~\ref{tab:swe_verified_subset_ids} lists the complete subset grouped by
repository. We report each repository prefix once and list
the corresponding numeric instance IDs; for example,
\texttt{astropy\_\_astropy} with instance ID 7671 denotes
\texttt{astropy\_\_astropy-7671}.

\begin{table}[h]
    \centering
    \caption{The 200-task SWE-bench Verified evaluation subset, grouped by repository.}
    \label{tab:swe_verified_subset_ids}
    \scriptsize
    \setlength{\tabcolsep}{4pt}
    \renewcommand{\arraystretch}{1.08}
    \begin{tabular}{@{}p{0.27\linewidth}p{0.68\linewidth}@{}}
        \toprule
        Repository & Instance IDs \\
        \midrule
        \texttt{astropy\_\_astropy} & 7671, 8707, 13579, 13977, 14365, 14369, 14995 \\
        \texttt{django\_\_django} & 9296, 10097, 10914, 10999, 11095, 11133, 11138, 11163, 11179, 11211, 11276, 11292, 11299, 11400, 11433, 11477, 11490, 11532, 11728, 11734, 11749, 11815, 11880, 11951, 11964, 11999, 12125, 12155, 12193, 12209, 12262, 12273, 12276, 12419, 12708, 13012, 13112, 13128, 13158, 13297, 13343, 13344, 13346, 13363, 13449, 13516, 13551, 13569, 13658, 13741, 13786, 13809, 13810, 13820, 13821, 13837, 13933, 14007, 14089, 14155, 14170, 14238, 14404, 14493, 14539, 14559, 14580, 14672, 14725, 14765, 14771, 14787, 14792, 14999, 15022, 15037, 15098, 15128, 15252, 15278, 15368, 15382, 15467, 15499, 15554, 15563, 15572, 15731, 15741, 15863, 16032, 16100, 16136, 16256, 16263, 16333, 16429, 16485, 16502, 16595, 16612, 16631, 16661, 16662, 16819, 16901 \\
        \texttt{matplotlib\_\_matplotlib} & 20676, 22719, 24149, 24627, 24970, 25311, 25775, 26342 \\
        \texttt{psf\_\_requests} & 1142, 1766, 1921, 2317, 5414 \\
        \texttt{pydata\_\_xarray} & 2905, 3151, 3305, 4094, 4356, 4629, 6461, 7229 \\
        \texttt{pylint-dev\_\_pylint} & 4604, 6903, 7277, 8898 \\
        \texttt{pytest-dev\_\_pytest} & 5631, 5840, 7205, 7236, 7521, 7982, 10051, 10081 \\
        \texttt{scikit-learn\_\_scikit-learn} & 9288, 10297, 11310, 13124, 13439, 14141, 14496, 15100 \\
        \texttt{sphinx-doc\_\_sphinx} & 7440, 7462, 7757, 7889, 7985, 8035, 8056, 8120, 8265, 8459, 8475, 8548, 9258, 9591, 9658, 10435, 10466 \\
        \texttt{sympy\_\_sympy} & 11618, 12096, 12419, 13091, 13372, 13480, 13551, 13757, 13852, 13974, 14248, 14976, 15349, 15875, 15976, 16450, 16597, 17318, 18211, 18698, 19346, 19495, 19637, 22714, 22914, 23262, 23534, 24066, 24539 \\
        \bottomrule
    \end{tabular}
\end{table}

\subsection{Training and Compute Details}
\label{app:training_compute_details}

\paragraph{Model and compute configurations.}
The policy and critic are initialized from the same backbone, with the critic
augmented by a scalar value head. 
All hyperparameter settings are identical for both the 30B and 106B configurations. Table~\ref{tab:training_compute} summarizes the training configurations and compute resources for both models. Table~\ref{tab:supo_compute} compares the training-resource
usage of \model{} and the GRPO baseline under GLM-4.5-Air(106B-A12B).
We additionally repeated \model{} training with two random seeds under the same configuration and observed similar performance across seeds.


\begin{table}[h]
    \centering
    \caption{Training configurations and compute resources for \model{}.}
    \label{tab:training_compute}
    \setlength{\tabcolsep}{5pt}
    \renewcommand{\arraystretch}{1.08}
    \begin{tabular}{@{}lp{0.68\linewidth}@{}}
        \toprule
        Item & Configuration \\
        \midrule

        \multicolumn{2}{@{}l}{\textbf{GLM-4.7-Flash (30B-A3B)}} \\
        \addlinespace[2pt]
        Model architecture &
        47 layers, hidden size 2,048, 20 attention heads \\
        MoE configuration &
        64 routed experts, top-4 routing, 1 shared expert \\
        Runtime / compute &
        180 training steps in 15 hours; 960 GPU-hours \\
        
        \midrule

        \multicolumn{2}{@{}l}{\textbf{GLM-4.5-Air (106B-A12B)}} \\
        \addlinespace[2pt]
        Model architecture &
        46 layers, hidden size 4,096, 96 attention heads \\
        MoE configuration &
        128 routed experts, top-8 routing, 1 shared expert \\
        Runtime / compute &
        100 training steps in 23 hours; 1,472 GPU-hours \\
        
        \midrule

        \multicolumn{2}{@{}l}{\textbf{Shared training configuration}} \\
        \addlinespace[2pt]
        Policy optimization &
        Adam, constant LR $2 \times 10^{-6}$,
        $\beta=(0.9,0.98)$, weight decay 0.1 \\
        Critic optimization &
        LR $3 \times 10^{-6}$, two critic updates per policy update \\
        Advantage estimation &
        $\gamma=1$, length-adaptive GAE with
        $\lambda=1-\frac{1}{\alpha l}$ and $\alpha=1.5$ \\
        KL &
        KL coefficient 0 \\
        Hardware &
        64 NVIDIA H800 GPUs: 32 shared by actor and critic,
        and 32 for rollouts \\
        \bottomrule
    \end{tabular}
\end{table}

\begin{table}[h]
    \centering
    \caption{Training-resource comparison between \model{} and the GRPO baseline.}
    \label{tab:supo_compute}
    \setlength{\tabcolsep}{4pt}
    \begin{tabular}{@{}lcccccc@{}}
        \toprule
        Method & \makecell{Total\\prompts} & \makecell{Batch size\\(segments)} & \makecell{Total\\rollouts} & \makecell{Generated\\tokens} & \makecell{Wall-clock\\time} & GPU-hours \\
        \midrule
        \model{} & 1,405 & 128 & 11,092 & 442M & 23 h & 1,472 \\
        SUPO & 1,405 & 128 & 11,484 & 448M & 25 h & 1,600 \\
        \bottomrule
    \end{tabular}
\end{table}

\FloatBarrier
\subsection{Compaction Prompt Templates}
\label{app:compaction_prompts}

We use fixed templates for summary generation and post-compaction resumption.
These correspond to $q_{\mathrm{sum}}$ and $u_{\mathrm{resume}}(S_t)$ in
Section~\ref{sec:trainable_compaction}. The original task instruction is explicitly included in the
summary-generation prompt, while the most recent $k$ assistant--tool-result
pairs are retained verbatim after compaction.

\begin{promptbox}[Summarization Prompt $q_{\mathrm{sum}}$]
Your job is to create a detailed summary of the conversation so far,
paying close attention to the user's explicit requests and the work
already completed.

\medskip
Original task:

\{ORIGINAL\_INSTRUCTION\}

\medskip
You are summarizing the earlier portion of the conversation only.
\{RECENT\_MESSAGE\_NOTE\} will remain in the conversation verbatim, so do
not repeat them in detail unless they are required for continuity.

\medskip
Do not assume you can read external files or use tools. Work only from
the conversation provided in this context.

\medskip
Before providing the summary, think through what matters most to
preserve. Write the response using exactly these tags:

\medskip
<analysis>
Brief internal reasoning about what to preserve.
</analysis>

<summary>
Your summary here.
</summary>

\medskip
Inside <summary>, organize the content so the next model can continue
without reading any external files. Include:

1. Primary Request and Intent

2. Key Technical Concepts

3. Files and Code Sections examined or changed

4. Errors encountered and how they were addressed

5. Problem Solving and important decisions

6. All user messages and explicit asks from the summarized portion

7. Pending Tasks

8. Current Work

9. Optional Next Step

\medskip
Be concrete. Preserve exact commands, outputs, code behavior, and error
text when those details matter for the work to continue. Do not output JSON.
\end{promptbox}

\begin{promptbox}[Post-Compaction Resume Prompt $u_{\mathrm{resume}}(S_t)$]
This session is being continued from a previous conversation that
ran out of context. The summary below covers the earlier portion
of the conversation.

\medskip
\{SUMMARY\_TEXT\}

\medskip
Recent messages are preserved verbatim below and should be treated
as the source of truth for the latest state.

\medskip
Continue the conversation from where it left off without asking any
further questions. Resume directly from this summary and the
preserved recent messages.
\end{promptbox}

\section{Analysis on Hyperparameter Sensitivity}
\label{app:compaction_sensitivity}

We evaluate the hyperparameter sensitivity of \model{} on the 200-task SWE-bench Verified subset using the trained \model{} checkpoint of GLM-4.5-Air (106B-A12B). We vary the context budget, maximum number of compactions, and retained history length at inference time. The context budget denotes the maximum working context length $C$, and $k$ denotes the number of recent agent--environment interaction turns retained after compaction.

\begin{table}[h]
    \centering
    \caption{Sensitivity to inference-time compaction settings on SWE-bench Verified. Results are Pass@1 accuracy (\%) reported as mean $\pm$ standard deviation over evaluation runs.}
    \label{tab:compaction_sensitivity}
    \setlength{\tabcolsep}{8pt}
    \begin{tabular}{@{}lcc@{}}
        \toprule
        Factor & Setting & SWE-bench Verified \\
        \midrule
        \multirow{3}{*}{Context budget}
            & 32k & $63.0 \pm 0.5$ \\
            & 64k & $65.5 \pm 1.7$ \\
            & 80k & $66.4 \pm 0.7$ \\
        \midrule
        \multirow{3}{*}{Maximum compactions}
            & 0 & $59.1 \pm 2.2$ \\
            & 1 & $65.3 \pm 1.5$ \\
            & 3 & $66.4 \pm 0.7$ \\
        \midrule
        \multirow{2}{*}{Retained history length}
            & $k=0$ & $66.0 \pm 0.5$ \\
            & $k=2$ & $66.4 \pm 0.7$ \\
        \bottomrule
    \end{tabular}
\end{table}

Table~\ref{tab:compaction_sensitivity} shows that \model{} maintains stable performance within practical hyperparameter ranges, indicating limited sensitivity to these settings. Removing the recent-history tail causes only a minor performance drop, suggesting that the learned summaries retain most task-relevant state.

\section{Summary Quality Analysis}
\label{app:summary_quality_analysis}

We evaluate summary quality on SWE-bench Verified trajectories where at least one of CompactionRL and RL trained without compaction, when evaluated with inference-time compaction, triggers at least one compaction. The analysis covers 79 samples from the 200-task SWE-bench Verified subset. When multiple compaction summaries are available within the same trajectory, we first average their scores. To reduce uncertainty from the LLM judge, each evaluation is repeated three times. As shown in Table~\ref{tab:summary_quality_judge}, \model{} obtains a higher average score than RL without compaction training, with an overall mean difference of $+0.22$ and a 95\% bootstrap confidence interval of $[0.020, 0.420]$.

\begin{table}[h]
    \centering
    \caption{Summary-quality evaluation of RL without compaction training and \model{}. Scores use a 0--5 scale.}
    \label{tab:summary_quality_judge}
    \setlength{\tabcolsep}{3pt}
    \resizebox{\linewidth}{!}{
    \begin{tabular}{@{}lp{0.47\linewidth}cc@{}}
        \toprule
        Dimension & Evaluation criterion & \makecell{RL (w/o\\compaction training)} & \model{} \\
        \midrule
        Requirement Fidelity & Whether the summary accurately preserves task requirements, constraints, and acceptance criteria & 3.67 & \textbf{4.00} \\
        Current-State Fidelity & Whether the summary accurately records the current repository, patch, test, and completion state & 3.28 & \textbf{3.53} \\
        Failure-History Fidelity & Whether the summary preserves key failures, diagnoses, and rejected approaches & 3.38 & \textbf{3.48} \\
        Next-Step Validity & Whether the pending work and next steps are correct, specific, executable, and non-misleading & 3.14 & \textbf{3.43} \\
        Continuation Usefulness & Whether the summary reduces redundant exploration caused by missing or misleading state and promotes new, verifiable progress & 3.05 & \textbf{3.15} \\
        Total & -- & 3.30 & \textbf{3.52} \\
        \bottomrule
    \end{tabular}
    }
\end{table}

The score distribution also shifts away from low-quality summaries: the fraction of scores in the 0--2 range decreases from 22.2\% for RL without compaction training to 11.9\% for \model{}. On the subset where both methods compact, \model{} also achieves a higher task success rate, 38.10\% versus 28.57\%, consistent with its higher summary-quality scores.

\FloatBarrier
\paragraph{Judge setup.}
\label{app:summary_judge_setup}
The judge uses \texttt{claude-opus-5} with \texttt{reasoning\_effort=medium}. The evaluation prompt is shown below.

\begin{promptbox}[Summary-Quality Judge Prompt]
  \footnotesize
  
  \textbf{Role.}
  You are a trajectory summary evaluator. Assess a compaction summary for
  \textbf{Faithfulness} and \textbf{Continuation Usefulness}.
  
  \medskip
  \textbf{The trace contains:}
  \begin{itemize}[leftmargin=1.4em, nosep]
      \item \textbf{Segment 1:} original prompt, pre-summary execution,
      summary prompt, and candidate summary.
      \item \textbf{Segment 2:} original prompt, candidate summary,
      and post-summary execution.
  \end{itemize}

  \medskip
  \textbf{Rules.}
  \begin{itemize}[leftmargin=1.4em, nosep]
      \item Evaluate whether the summary accurately preserves
      continuation-critical requirements, state, failures, blockers,
      and next steps.
      \item Separate summary quality from continuation-policy, execution,
      and environment quality.
      \item Cite concrete trace evidence for every score.
      \item Judge relevant completeness and information density,
      not length or detail count.
      \item The final two pre-summary assistant/tool-result pairs form
      the retained tail. Do not penalize facts already clear there.
      \item Use post-summary execution only for continuation usefulness,
      and penalize only summary-caused recovery errors, repetition,
      or detours.
  \end{itemize}
  
  \medskip
  \textbf{Scores.}
  Use 0--5: 5 nearly perfect; 4 high quality; 3 usable with clear defects;
  2 important defects; 1 core state missing or wrong; 0 unusable;
  ``N/A'' if there is insufficient evidence.
  
  \medskip
  \textbf{Dimensions.}
  \begin{itemize}[leftmargin=1.4em, nosep]
      \item \textbf{F1 Requirement Fidelity:} Preserves required constraints
      and acceptance criteria.
      \item \textbf{F2 Current-State Fidelity:} Accurately preserves repository,
      file, patch, test, environment, and completion state.
      \item \textbf{F3 Failure-History Fidelity:} Preserves failures, diagnoses,
      rejected hypotheses, and costly ineffective approaches.
      \item \textbf{F4 Next-Step Validity:} Gives correct, specific, executable
      next steps consistent with the current state.
      \item \textbf{U Continuation Efficiency and Progress:} Helps recover state,
      avoid summary-induced repetition or detours, and make verifiable progress.
  \end{itemize}
  
  \medskip
  \textbf{Output.}
  Output exactly one valid JSON object and no other text.
  \begin{Verbatim}[fontsize=\scriptsize,breaklines=true]
  {
    "faithfulness": {
      "F1_requirement_fidelity":
        {"score": 0, "evidence": [], "rationale": "", "key_issues": []},
      "F2_current_state_fidelity":
        {"score": 0, "evidence": [], "rationale": "", "key_issues": []},
      "F3_failure_history_fidelity":
        {"score": 0, "evidence": [], "rationale": "", "key_issues": []},
      "F4_next_step_validity":
        {"score": 0, "evidence": [], "rationale": "", "key_issues": []}
    },
    "usefulness": {
      "U_continuation_efficiency_and_progress":
        {"score": 0, "evidence": [], "rationale": "", "key_issues": []}
    }
  }
  \end{Verbatim}
  
  \medskip
  \textbf{Trace.}
  \begin{Verbatim}[fontsize=\scriptsize,breaklines=true]
  <segment_1>
  {{SYSTEM_AND_USER_PROMPTS}}{{PRE_SUMMARY_EXECUTION}}{{SUMMARY_PROMPT}}
  <candidate_summary>{{CANDIDATE_SUMMARY}}</candidate_summary>
  </segment_1>
  <segment_2>
  {{INITIAL_PROMPT_AND_CANDIDATE_SUMMARY}}{{POST_SUMMARY_EXECUTION}}
  </segment_2>
  \end{Verbatim}

  \end{promptbox}

\paragraph{Failure modes.}
\label{app:summary_failure_modes}
We identify three recurring failure modes in lower-quality summaries: (1) omitted or misdiagnosed failures, where the summary records that tests failed but preserves an incorrect diagnosis; (2) stale or misrepresented repository states, where temporary debugging choices are mistaken for the final solution; and (3) premature completion claims, where passing tests are treated as sufficient even though the implementation still violates an explicit task requirement. In a representative comparison, RL (w/o compaction training) treats an earlier database failure as the current test status despite a later successful test run, causing the post-compaction agent to continue debugging an obsolete issue and ultimately fail. By contrast, \model{} preserves a more detailed account of the execution state, including which file-writing attempts failed, how these failures corrupted the source file, and why the task remained incomplete. These details guide the post-compaction agent to recover the file, complete the required modification, and verify the solution.

\end{document}